\documentclass[10pt,twocolumn,letterpaper]{article}

\usepackage{iccv}
\usepackage{times}
\usepackage{epsfig}
\usepackage{graphicx}
\usepackage{amsmath}
\usepackage{amssymb}

\usepackage{booktabs}
\usepackage{multirow}
\usepackage[table,xcdraw]{xcolor}

\usepackage[pagebackref=true,breaklinks=true,letterpaper=true,colorlinks,bookmarks=false]{hyperref}

\iccvfinalcopy 

\def\iccvPaperID{****} 

\ificcvfinal\fi

\begin{document}

\title{Enhancing Building Semantic Segmentation Accuracy with Super Resolution and Deep Learning: Investigating the Impact of Spatial Resolution on Various Datasets}


\author{
Zhiling Guo$^{\textit{1,2}}$,
Xiaodan Shi$^{\textit{2}}$,
Haoran Zhang$^{\textit{2}}$,
Dou Huang$^{\textit{2}}$,
Xiaoya Song$^{\textit{3}}$, \\
Jinyue Yan$^{\textit{1}}$,
Ryosuke Shibasaki$^{\textit{2}}$
\\
$^{\textit{1}}$Department of Building Environment and Energy Engineering, \\ The Hong Kong Polytechnic University, Kowloon, Hong Kong, China
\\ $^{\textit{2}}$Center for Spatial Information Science, The University of Tokyo, Kashiwa, Japan
\\ $^{\textit{3}}$School of Architecture, Harbin Institute of Technology, Harbin, China
}



\maketitle
\ificcvfinal\thispagestyle{empty}\fi

\begin{abstract}
The development of remote sensing and deep learning techniques has enabled building semantic segmentation with high accuracy and efficiency. Despite their success in different tasks, the discussions on the impact of spatial resolution on deep learning based building semantic segmentation are quite inadequate, which makes choosing a higher cost-effective data source a big challenge. To address the issue mentioned above, in this study, we create remote sensing images among three study areas into multiple spatial resolutions by super-resolution and down-sampling. After that, two representative deep learning architectures: UNet and FPN, are selected for model training and testing. The experimental results obtained from three cities with two deep learning models indicate that the spatial resolution greatly influences building segmentation results, and with a better cost-effectiveness around 0.3m, which we believe will be an important insight for data selection and preparation.
\end{abstract}

\section{Introduction}
\label{sec:intro}
Buildings semantic segmentation via remote sensing imagery has become an important research topic in recent years \cite{guo2019super}. With the rapid development of data acquisition systems and machine learning, the ever-expanding choices of datasets with very high resolution (VHR) \cite{xia2018dota} and deep learning methods \cite{goodfellow2016deep} expand the opportunities for researchers to conduct more accurate analysis. 

\begin{figure}[t]
\centering
\includegraphics[width=\columnwidth]{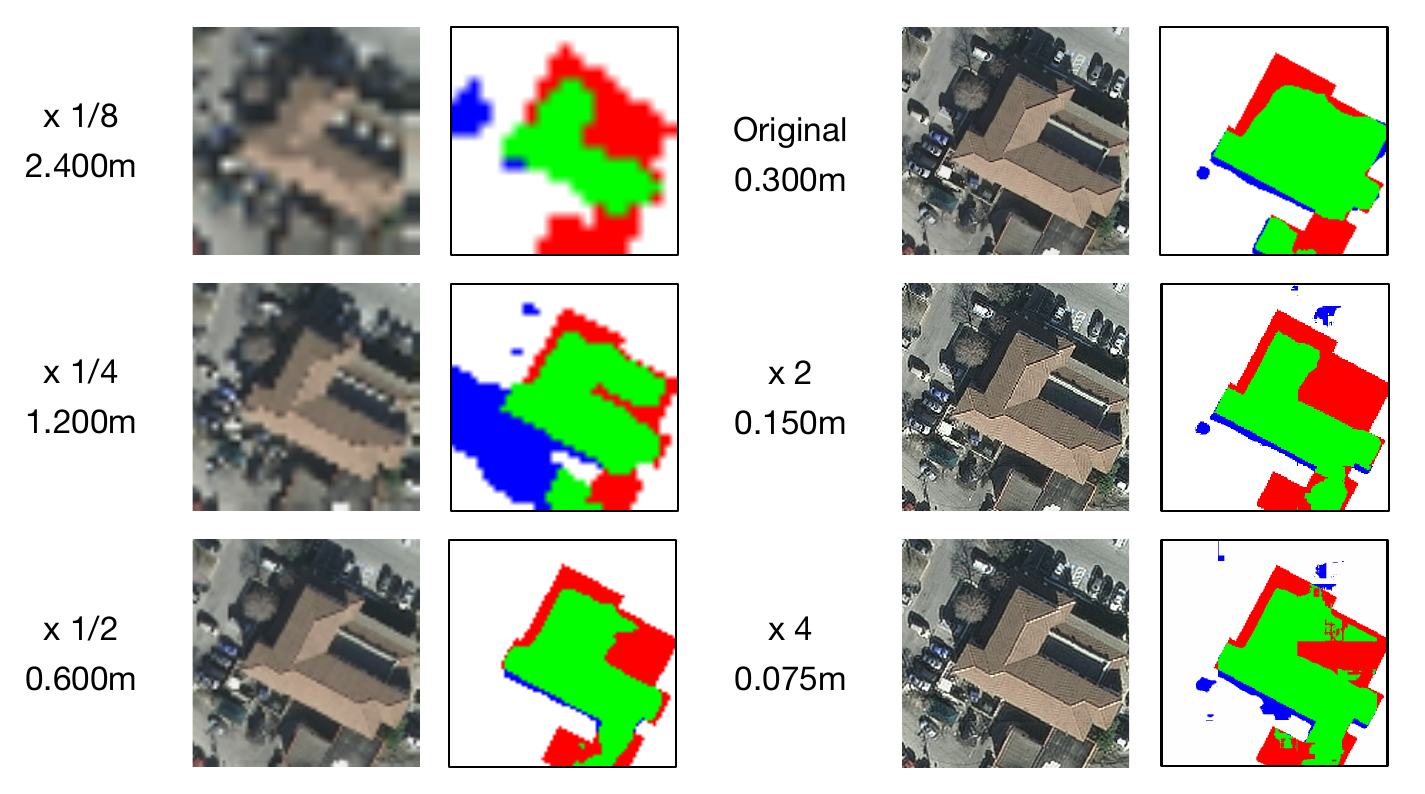}
\caption{The impact of spatial resolution on deep-learning based building semantic segmentation. The different colors: green, red, blue, and white, are used to indicate the true positive (tp), false negative (fn), false positive (fp), and true negative (tn) pixels in segmentation results, respectively.}
\label{fig:teaser}
\end{figure}

Although VHR imagery would express finer information contents of the landscape, it requires higher cost, longer processing time, and bigger storage space. Thus, the evaluation of the technical and economic trade-offs associated with using different resolution imagery is essential. The previous scholars have studied the impact of resolution in plant species \cite{roth2015impact}, land use \cite{liu2020impact}, and water \cite{fisher2018impact} pattern recognition based on coarser-resolution or conventional machine learning methods. In this study, we investigate the impact of spatial resolution for building semantic segmentation via VHR imagery and deep learning methods, as shown in figure \ref{fig:teaser}.

To compare the segmentation accuracy under different resolutions, we created remote sensing imagery in a specific area with resolutions from 0.075m to 2.4m by super-resolution (SR) \cite{glasner2009super} and down-sampling processing. The experimental results obtained from three different study areas via two deep learning models reveal that the finer the spatial resolution may not be the best in building semantic segmentation tasks, and the relatively low-cost imagery would be sufficient in many study cases. Thus, choosing a cost-effective spatial resolution for different scenarios is worth discussing.


The main contributions of this study can be highlighted as two folds. First, to the best of our knowledge, it is the first investigation for the impact of spatial-resolution on deep learning-based building semantic segmentation. Second, the resolution is not the higher the better for segmentation accuracy. According to our dataset, a resolution around 0.3m is better for cost-effectiveness, which enables researchers and developers to conduct their research efficiently.

\section{Data}\label{Data}
We analyzed the impact of spatial resolution for building semantic segmentation over three representative study areas: Austin, Christchurch, and Tokyo. The original resolutions of the datasets mentioned above are about 0.075m, 0.150m, and 0.300m, respectively.

\begin{figure}[h]
\centering
\includegraphics[width=\columnwidth]{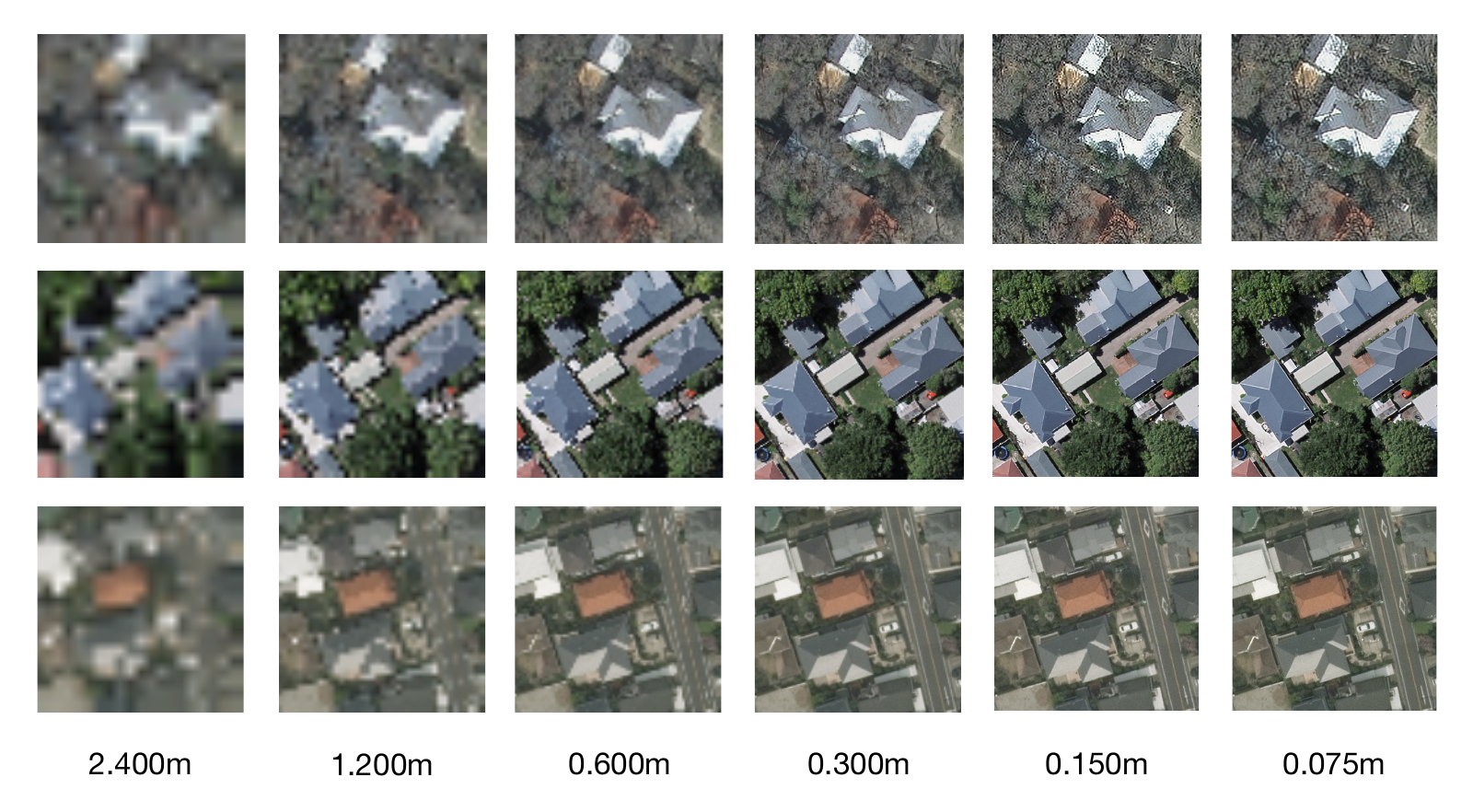}
\caption{Building illustrated in remote sensing images with different resolutions. Samples from Austin, Christchurch, and Tokyo are shown in the first, second, and third rows. The original resolution was up- and down-scaled by SR and down-sampling, respectively.}
\label{fig:reso_compare}
\end{figure}

\section{Methods}\label{methods}
The variation of spatial resolution will lead to differences in semantic segmentation results. At first, we resampled the imagery to a total of six pixel scales according to the spatial resolution range of most VHR images in data preprocessing, as shown in figure \ref{fig:reso_compare}. After that, two representative semantic segmentation models are applied for building semantic segmentation. Finally, the comparison is conducted based on four assessment criteria.

\subsection{Preprocessing}\label{Preprocessing}
Compared with upscaling low-resolution imagery to HR space using a single filter such as bicubic interpolation, SR could increase the image resolution while providing finer spatial details than those captured by the original acquisition sensors. In this study, one of the typical deep learning SR models: ESPCN \cite{shi2016real} is utilized to perform SR. In terms of the resample to lower-resolution, the pixel aggregate method is adopted. After that, six pixel scales in 0.075m, 0.150m, 0.300m, 0.600m, 1.200m, 2.400m can be generated.

\subsection{Semantic Segmentation}\label{semantic segmentation}
As the representative deep learning models, in this study, we propose to adopt UNet \cite{ronneberger2015u} and FPN \cite{lin2017feature}  to conduct the building semantic segmentation and investigate the impact of spatial Resolution in results. In general, Unet applies multiple skip connections between upper and downer layers, while FPN obtains features in bottom-up and top-down pathways. Both models have shown the high feasibility and robustness in many segmentation tasks. It should be noted, the data augmentation methods are adopted without random scaling in training, and a model trained by a specific area and resolution is applied to test the corresponding area and resolution for a fair comparison. 

\begin{figure}[!hbt]
\centering
\includegraphics[width=\columnwidth]{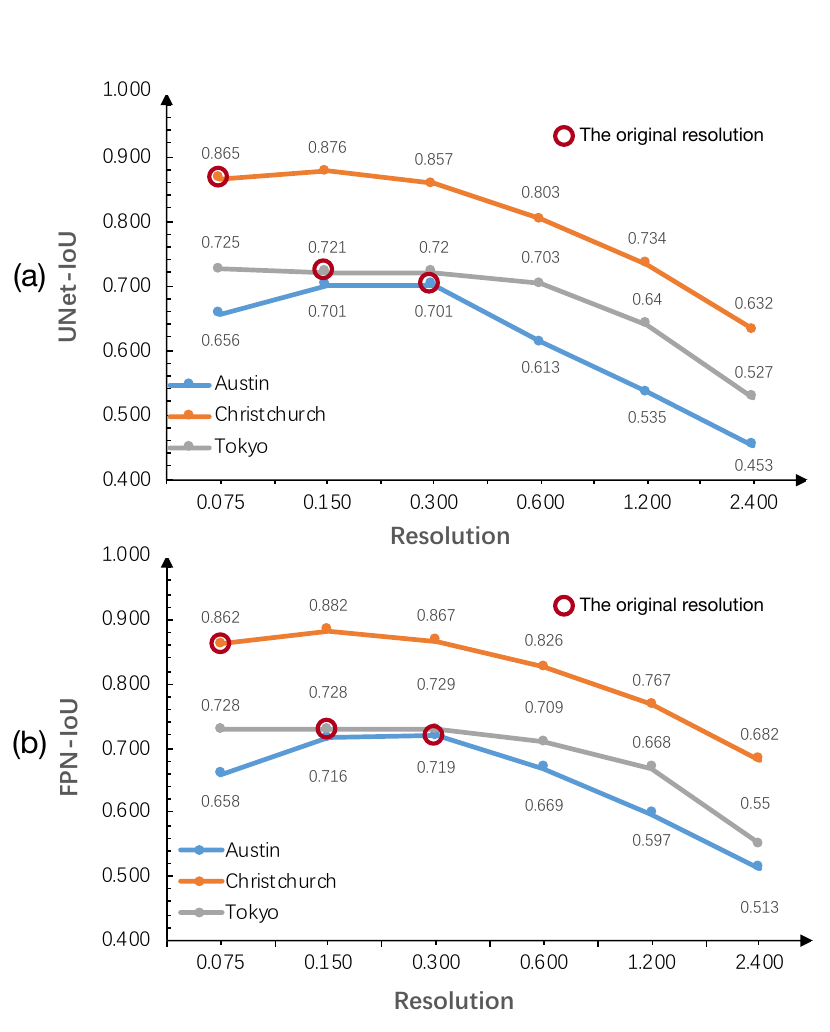}
\caption{The impact of spatial resolution on building semantic segmentation via (a) UNet and (b) FPN in Austin, Christchurch, and Tokyo, respectively.}
\label{fig:organized}
\end{figure}

\section{Results and Discussions}\label{results}

After testing, we generated segmentation results in three cities with different resolutions by two deep learning architectures. Figure \ref{fig:organized} illustrates the impact of spatial resolution on deep learning-based building semantic segmentation, and the detailed quantitative results in IoU can be found in Table \ref{table:quantitative result}. It can be seen that resolution significantly influences the segmentation results, although images in some resolutions are generated by resampling methods. With the decrease of spatial resolution, in the beginning, the IoU increases slightly in Austin and is stable in both Christchurch and Tokyo. After a certain threshold of 0.300m, the IoU drops rapidly in all study areas. Importantly, both UNet and FPN show a similar tendency. This makes sense, as building features have specific physical size, and the spatial resolution is significantly finer than the certain threshold, which may not help the segmentation performance while providing redundant information. Therefore, the spatial resolution should reach a certain threshold to achieve decent accuracy, and the excessively pursue of finer resolution than the threshold is no need in many cases. Such a trade-off should be involved while selecting an appropriate data source. The experimental results obtained from three cities with two deep learning models demonstrate that the resolution is not the higher the better, and 0.3m resolution would be a better cost-effective choice for data selection and preparation in building semantic segmentation tasks.

\label{ssec:quantitative result}
\begin{table}[!hbt]
\caption{Quantitative comparison of different models. {\color[HTML]{FE0000} Red} refers to the result from original resolution, and \textbf{bold} represents the best result.\\}
\resizebox{0.47\textwidth}{!}{
\begin{tabular}{ccccccc}
\hline
                             & \multicolumn{2}{c}{Austin}                                                    & \multicolumn{2}{c}{Christchurch}                            & \multicolumn{2}{c}{Tokyo}                                   \\ \cline{2-7} 
\multirow{-2}{*}{Resolution} & UNet                                  & FPN                                   & UNet                         & FPN                          & UNet                         & FPN                          \\ \hline
0.075                        & 0.656                                 & 0.658                                 & {\color[HTML]{FE0000} 0.865} & {\color[HTML]{FE0000} 0.862} & \textbf{0.725}               & 0.728                        \\
0.150                        & \textbf{0.701}                        & 0.716                                 & \textbf{0.876}               & \textbf{0.882}               & {\color[HTML]{FE0000} 0.721} & {\color[HTML]{FE0000} 0.728} \\
0.300                        & {\color[HTML]{FE0000} \textbf{0.701}} & {\color[HTML]{FE0000} \textbf{0.719}} & 0.857                        & 0.867                        & 0.720                        & \textbf{0.729}               \\
0.600                        & 0.613                                 & 0.669                                 & 0.803                        & 0.826                        & 0.703                        & 0.709                        \\
1.200                        & 0.535                                 & 0.597                                 & 0.734                        & 0.767                        & 0.640                        & 0.668                        \\
2.400                        & 0.453                                 & 0.513                                 & 0.632                        & 0.682                        & 0.527                        & 0.550                        \\ \hline

\end{tabular}}
\label{table:quantitative result}
\end{table}

\section{Conclusion}
In this study, we have investigated the impact of spatial resolution on deep learning-based building semantic segmentation and demonstrated the effectiveness of super resolution techniques in enhancing segmentation accuracy. Our results suggest that spatial resolution plays a critical role in the accuracy and generalization capability of deep learning models for building semantic segmentation, and that super resolution techniques can help to overcome the limitations of low-resolution data. 

To further advance this line of research, future work could extend our empirical evaluation to other deep learning models, study areas, and data sources.


\section{Acknowledgement}
We are grateful for the support and funding provided by the JSPS 21K14261 grant.



{\small
\bibliographystyle{ieee_fullname}
\bibliography{egpaper_final}
}

\end{document}